\documentclass{article}
\usepackage{arxiv}
\usepackage[T1]{fontenc}
\usepackage{graphicx}
\usepackage{subcaption}
\usepackage{booktabs}
\usepackage{wrapfig}
\usepackage[table]{xcolor}
\definecolor{rowgray}{gray}{0.85}
\definecolor{rowblue}{RGB}{218,232,252}
\usepackage{multirow}
\usepackage{adjustbox}

\usepackage[colorlinks,citecolor=blue]{hyperref}
\usepackage[table]{xcolor} 
\definecolor{rowgray}{gray}{0.85}
\definecolor{rowblue}{RGB}{218,232,252}
\definecolor{mygray}{gray}{0.9}
\definecolor{myblue}{rgb}{0.88, 0.94, 0.99}
\usepackage{booktabs}

\usepackage{orcidlink}
\usepackage{amsmath}
\usepackage{amssymb}
\usepackage{cleveref}

\begin{document}

\title{PathSelect: Sequential Token Selection for Whole Slide Pathology}

\author{
  Jingzhi Chen\footnotemark[1] \\
  Shenzhen University of Advanced Technology \\
\And
  Landi He\footnotemark[1] \\
  Shenzhen University of Advanced Technology \\
  \And
  Zehong Chen \\
  Shenzhen University of Advanced Technology \\
  \And
  Peihang Wu \\
  Shenzhen University of Advanced Technology \\
  \And
  Lijian Xu\footnotemark[2] \\
  Shenzhen University of Advanced Technology \\
}
\footnotetext[1]{These authors contributed equally to this work.}
\footnotetext[2]{Corresponding author.}

\maketitle

\begin{abstract}
Gigapixel Whole-Slide Images (WSIs) present a fundamental computational bottleneck for vision-language models (VLMs) due to extreme sequence lengths. Existing approaches predominantly rely on spatial sampling or training-free pruning, which risk diluting weak but informative signals, leading to the loss of critical diagnostic evidence due to the spatially diffuse nature of pathological cues. We reformulate WSI token pruning as a \textbf{sequential selection process}, enabling the model to autonomously learn an optimal routing strategy rather than relying on static heuristics. In this work, we propose a decoupled routing framework integrated as an active plugin into the fully pre-trained SlideChat base model, leaving both the slide encoder and large language model (LLM) backbones strictly frozen. To provide continuous gradients for the non-differentiable pruning operation during training, we introduce \textbf{PathSelect}. PathSelect employs a variance-preserving (VP) noise gate to modulate each patch's information flow via a differentiable Soft Top-$K$ operator, paired with a diagonal-attention Denoiser that recovers the perturbed representations without semantic leakage. At inference, the PathSelect module is entirely detached. Relying solely on the trained Scorer, a deterministic Hard Top-$K$ operator executes adaptive, data-dependent trajectory termination, significantly accelerating downstream generative processing with exceptionally low sequential token selection latency. Driven by an empirical average of only $44.86$ tokens under a maximum constraint of $K=128$, our framework achieves $74.00\%$ overall accuracy on SlideBench (TCGA), representing an approximate $36.6\times$ spatial token reduction relative to the uncompressed baseline average while consistently outperforming sampling-based counterparts. It further demonstrates competitive zero-shot generalization on SlideBench (BCNB) and WSI-VQA*. By mitigating the visual context bottleneck via a lightweight single-GPU single-stage training protocol, this work establishes an exceptionally efficient paradigm for end-to-end gigapixel WSI reasoning.
\end{abstract}


\section{Introduction}
\label{sec:intro}

Vision-language models (VLMs) \cite{young2026scalar,xu2026unified,xu2024foundation,young2026xrayclaw} have catalyzed a paradigm shift in computational pathology, enabling interactive diagnostic reasoning and slide-level question answering. However, scaling these models to gigapixel Whole-Slide Images (WSIs) presents a fundamental computational bottleneck. A single WSI typically generates over $10^5$ patches \cite{campanella2019clinical}, whose end-to-end processing amplifies the quadratic self-attention complexity and memory overhead during the large language model (LLM) prefill stage, rendering dense visual ingestion intractable  \cite{zheng2022graph,pinckaers2020streaming,chen2026tc}.

To mitigate this sequence explosion, prevailing paradigms resort to training-free visual token reduction. Spatial sampling strategies restrict the input to a fixed context window by uniformly or randomly discarding most patches \cite{chen2022scaling}. Training-free pruning methods rely on proxy metrics such as attention magnitude or visual similarity \cite{rao2021dynamicvit,bolya2022tome} to filter tokens before generative reasoning. While computationally efficient, these filters treat compression merely as identifying background noise, overlooking the extreme sparsity of pathology data where decisive diagnostic evidence, such as micro-metastases, often occupies less than 1\% of the slide.

\begin{figure*}[t]
\centering
\includegraphics[width=0.999\linewidth]{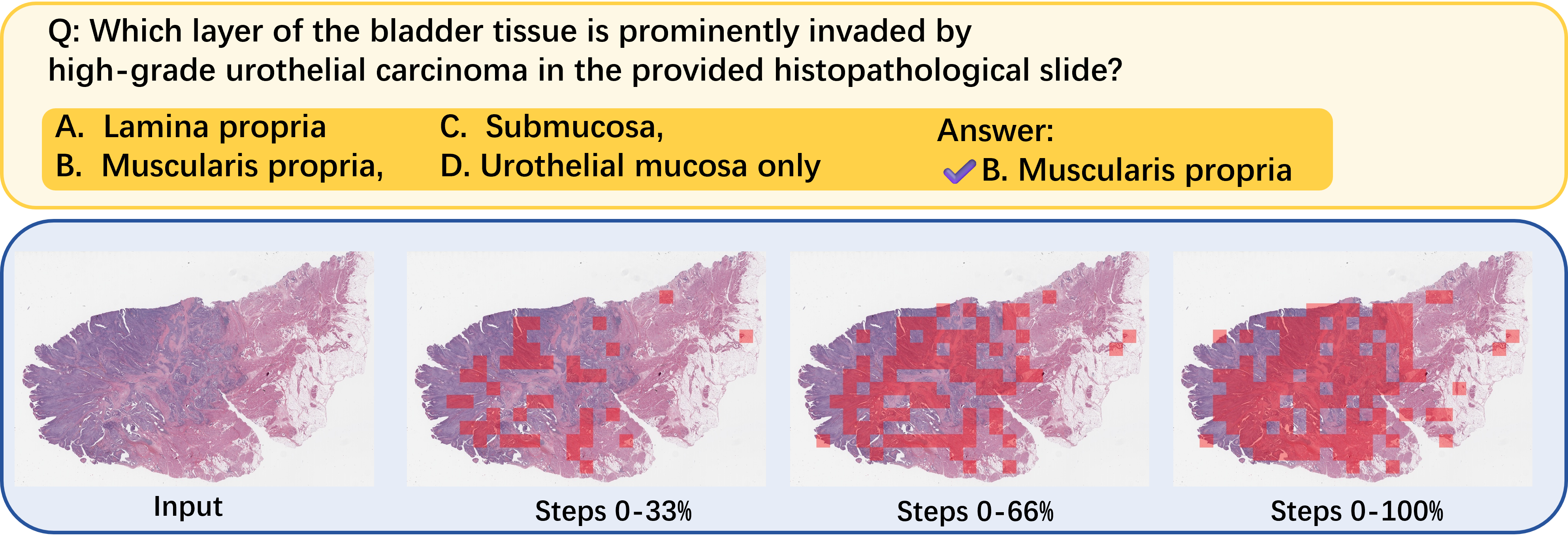}
\caption{Illustration of text-conditioned sequential token selection in PathSelect. The top panel presents a histology-based visual question-answering example regarding the invasion depth of high-grade urothelial carcinoma. The bottom panel sequentially visualizes the dynamic selection process across different decoding steps (Steps 0--33\%, 0--66\%, and 0--100\%), demonstrating how the model adaptively routes and concentrates on sparse, clinically decisive tumor regions from the raw input whole-slide image.}
\label{illustration}
\end{figure*}

To prevent information loss, recent efforts have shifted toward learnable token selection, exploring semantic slot aggregation for compact visual representations \cite{he2026beyond,he2026autoselect,he2026stepwise,chen2026learnable}. Yet, routing only the most informative tokens to the LLM inherently requires discrete, hard-gating decisions such as Hard Top-$K$ \cite{kool2019stochastic}, which are fundamentally non-differentiable \cite{bengio2013estimating}. Existing frameworks circumvent this by relying on surrogate objectives, external bounding-box annotations, or intrusive routing layers deep within transformer blocks \cite{takezoe2026learnpruner}, disrupting pre-trained language priors and hindering end-to-end multimodal alignment (see \Cref{illustration}).

We address this bottleneck by reconceptualizing WSI token pruning as a \textbf{data-dependent sequential selection process}. Instead of static, training-free spatial discard, we enable the model to autonomously learn an optimal token pruning strategy via continuous optimization. We propose a decoupled routing framework seamlessly integrated into the fully pre-trained SlideChat \cite{chen2025slidechat} baseline. Our PathSelect sequential selection module is directly plugged between the frozen visual encoder and the frozen LLM backbone. Guided by a Soft Top-$K$ operator, PathSelect employs a continuous variance-preserving (VP) noise gate to modulate each patch's information flow, creating continuous gradients without removing tokens during the training phase. A diagonal-attention Denoiser subsequently repairs latent distribution shifts while strictly preventing spatial information leakage. At inference, the auxiliary training components (i.e., the Noise Gate and Denoiser) are completely bypassed; a deterministic Hard Top-$K$ operator executes input-adaptive token retention based on the trained lightweight Scorer, introducing minimal computational latency while freeing the downstream LLM from dense context prefilling overheads.

Our contributions are summarized as follows:
\begin{itemize}
    \item We reformulate WSI token pruning from training-free heuristics into a learnable \textbf{sequential selection paradigm}, optimized end-to-end through continuous gradients without auxiliary task losses or architectural intrusions into the frozen LLM.
    \item We introduce a decoupled routing mechanism. To bridge the optimization gap of discrete sequential decision-making, we design PathSelect, a training plugin pairing a VP noise gate with diagonal-attention denoising to establish continuous gradients.
    \item Empirical evaluations validate our framework under an adaptive selection regime using a single NVIDIA A6000 GPU and a single-stage protocol. Under a maximum input constraint of $K=128$, our approach yields an empirical average of only $44.86$ tokens per slide, achieving a $36.6\times$ spatial token reduction relative to the dense SlideChat average. Under this compact regime, our framework yields $74.00\%$ overall accuracy on SlideBench (TCGA), with competitive zero-shot generalization on SlideBench (BCNB) and WSI-VQA*.
\end{itemize}

\section{Related Work}
\label{sec:related-work}

\subsection{Vision-Language Models for Gigapixel Pathology}
The integration of Multimodal Large Language Models (MLLMs) into gigapixel Whole Slide Image (WSI) analysis has established a new paradigm for cross-modal diagnostic reasoning \cite{yang2025one,young2026fewer}, typically built upon weakly supervised multiple instance learning (MIL) formulations. Early MIL frameworks introduced attention-based instance pooling \cite{ilse2018attention,lu2021data} to aggregate patch-level embeddings but generally treated instances independently. To capture long-range tissue structures, recent architectures have incorporated Transformer baselines \cite{shao2021transmil} alongside advanced variants leveraging mixture-of-experts \cite{hashimoto2024multimodal,wu2025learning}, dynamic fusion \cite{cao2023multi}, and slide-level multi-stage supervisions \cite{tang2024feature,tang2026revisiting}. 
Pathology-specific foundation models \cite{chen2024towards,xu2024whole,wu2026multimodal,xu2024medvilam}, including vision-language \cite{huang2023visual,lu2024visual} and multimodal variants \cite{sun2025cpath}, now encode raw patches as high-dimensional morphological tokens. Other representation learning strategies \cite{yang2024segmentation,yang2023geometry,feng2026efficient} have also been explored in medical imaging contexts.
 
Despite their empirical success, a single WSI typically generates hundreds of thousands of visual instances, precipitating an acute context window crisis when interfaced with MLLMs due to the quadratic complexity of global self-attention \cite{campanella2019clinical,chen2022scaling}. Conventional engineering remedies—such as sequence truncation, downsampling, or multi-stage localized slicing \cite{chen2022scaling}—mitigate computational scaling at the critical expense of severing global spatial topology and discarding fine-grained diagnostic clues. Crucially, existing visual feature reduction pipelines operate asynchronously from textual context encoding, causing a severe representation disconnect between downstream multimodal instruction interpretation and upstream region-of-interest selection (\textbf{Pain Point 1}). Furthermore, they uniformly enforce an instance-agnostic, static token retention budget $K$ across the entire cohort \cite{shao2025tokens}. This rigid constraint introduces a severe optimization mismatch between computational efficiency and diagnostic sensitivity (\textbf{Pain Point 3}): a inflated budget $K$ squanders excessive compute on redundant, low-entropy normal tissues, whereas a universally tightened $K$ risks catastrophic diagnostic omission on complex slides harboring minute, highly localized malignant lesions such as micro-metastases.

\subsection{Visual Token Compression in Gigapixel Pathology}
To resolve the sequence explosion, contemporary literature adapts token reduction techniques from natural computer vision, which generally branch into training-free heuristic selection \cite{bolya2022tome,dong2025mmtok,zhang2026beyond} and trainable pruning layers \cite{rao2021dynamicvit,huang2025dynamic,wu2026hidrop}. Training-free approaches rank visual tokens using closed-form proxy metrics derived from frozen attention statistics or static cross-modal text alignments \cite{yang2025visionzip,shao2025tokens}. While computationally efficient, these heuristics are fundamentally bottlenecked by inherent attention bias and lack the expressive capacity required to handle the dense semantic variations characteristic of pathological tissues.~

Concurrently, trainable token selection methods attempt to learn end-to-end routing policies by inserting parameterized gating links. However, because hard feature retention rules (e.g., top-$K$ selection) are intrinsically non-differentiable \cite{bengio2013estimating,jang2016categorical}, these models typically mandate discrete, multi-stage pipelines where feature extraction and selection are isolated into independent steps. This fragmented training scheme hinders joint parameter tuning, rendering the selection layer incapable of leveraging direct loss feedback propagated from downstream vision-language tasks. As a consequence, under the extreme spatial sparsity and weak slide-level supervision typical of WSI datasets, such uncoupled selection boundaries easily fall into sub-optimal local minima, causing the model to misidentify rare, clinically decisive tumor features as uninformative background and irrevocably discard them.

Beyond the multi-stage training limitation, a more pervasive flaw of current pruners lies in their underlying \textit{independence assumption}: each pathological patch is evaluated in absolute isolation \cite{shao2025tokens}. This structural formulation entirely ignores the spatial contiguity and dynamic marginal utility inherent to high-dimensional histological features (\textbf{Pain Point 2}). Within a coherent tissue architecture, once a specific tumor-associated patch is admitted into the retained subset, the marginal information gain of its immediate geometric neighbors drops sharply due to severe semantic redundancy. Conversely, the relative marginal contribution of distant but complementary semantic landmarks—such as distant tertiary lymphoid structures (TLS) or tumor-infiltrating lymphocytes (TILs)—increases significantly.

To decouple from the independent, one-shot top-$K$ selection paradigm \cite{hu2025loc,wang2026wsisum,guo2025focus}, our framework reformulates diagnostic token selection as an autoregressive sequential decision process. Driven by a pointer decoder, the policy dynamically updates candidate scores conditioned on the trajectory of prior selections, thereby naturally penalizing spatial redundancy while actively searching for complementary tissue structures. Combined with a continuous, variance-preserving noise-gating interface and a specialized length penalty objective, our approach achieves seamless end-to-end training while enabling fully input-adaptive token retention.

\section{Methodology}
\label{sec:methodology}

\begin{figure*}[t]
\centering
\includegraphics[width=0.99\linewidth]{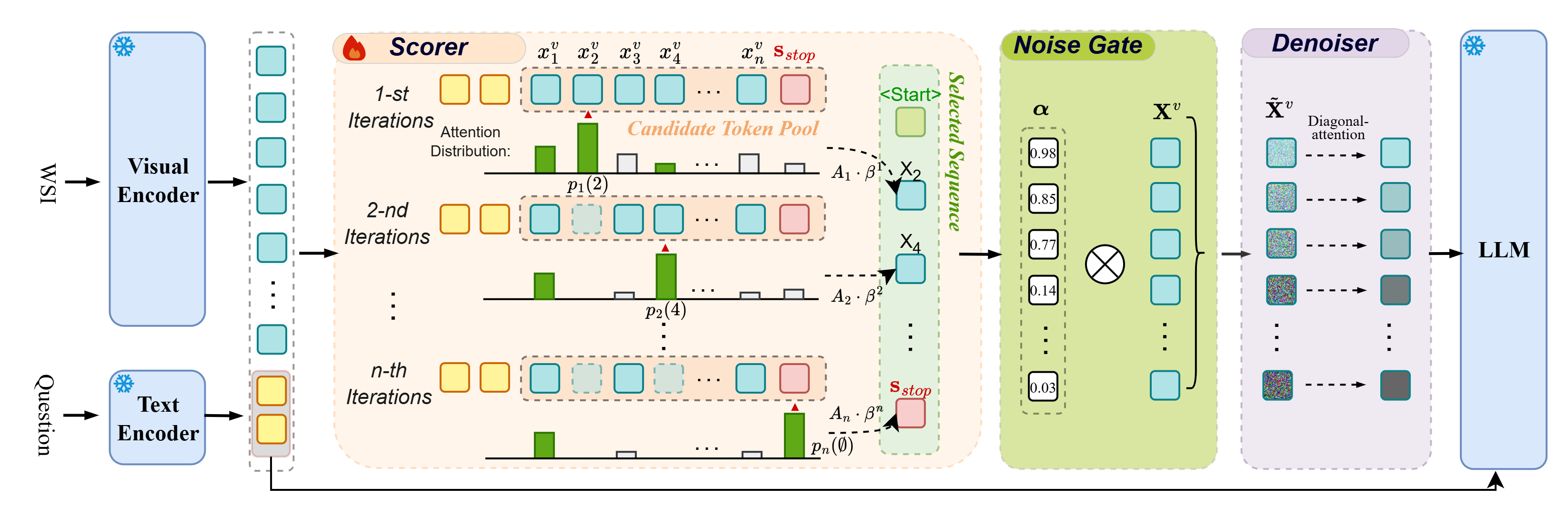}
\caption{\textbf{Overview of our framework.} 
Frozen encoders extract visual and textual features. 
\textbf{Autoregressive Selection Loop}: The joint cross-modal contextual representations are organized into a candidate token pool augmented with a virtual stop token $\mathbf{s}_{stop}$. Driven by a trainable conditional Scorer network, the pointer decoding trajectory sequentially selects prominent diagnostic regions step-by-step while calculating alive-weighted routing probabilities until the stop action $\emptyset$ is triggered. 
\textbf{Differentiable Training Path}: The continuous selection score vector $\boldsymbol{\alpha}$ is utilized by a VP Noise Gate to continuously perturb unselected background regions with isotropic Gaussian noise while safeguarding essential diagnostic semantics, followed by a diagonally-masked Denoiser block to realign the perturbed latent representations $\tilde{\mathbf{X}}^v$ before long-context language decoding. 
\textbf{Deterministic Inference Path}: The Noise Gate and Denoiser modules are detached, and the hard token subset is directly forwarded to the LLM for highly efficient downstream inference. Fire and snowflake icons represent learnable and frozen parameters, respectively.}
\label{fig:framework}
\end{figure*}

\subsection{System Formulation and Pipeline}
The comprehensive workflow of our proposed text-conditioned, end-to-end differentiable visual token selection paradigm is illustrated in Fig.~\ref{fig:framework}. Given a gigapixel WSI, we first discretize the tissue region into a sequence of $N$ non-overlapping patches and employ a frozen visual encoder to extract raw patch-level embeddings, yielding the initial visual feature matrix $\mathbf{X}^v = [\mathbf{x}_{1}^v, \mathbf{x}_{2}^v, \dots, \mathbf{x}_{N}^v]^\top \in \mathbb{R}^{N \times D}$. Concurrently, the user textual instruction is processed by a text encoder to generate word embeddings $\mathbf{E}^{txt} \in \mathbb{R}^{L_{txt} \times D_{txt}}$. 

To bridge the operational disconnect between upstream feature reduction and downstream textual understanding, we reformulate token compression as a conditional sequential decision process. Rather than relying on independent, static top-$K$ ranking frameworks that cannot capture dynamic marginal utility, the token selection policy is cast as an autoregressive formulation:
\begin{equation}
P(\mathcal{S}|\mathbf{X}^v) = \prod_{t=1}^{T} \pi_{\theta}(a_{t}|a_{<t}, \mathbf{X}^v),
\label{eq:autoregressive_policy}
\end{equation}
where $a_{t} \in \{1, \dots, N\} \cup \{\emptyset\}$ represents the action executed at step $t$, and $\emptyset$ denotes a learned stop action. As depicted in the Scorer module of Fig.~\ref{fig:framework}, the trajectory terminates at the first instance where $a_{t} = \emptyset$ (denoted as $T^*$), enabling the model to adaptively determine the retention length $K = T^* - 1$ on a per-sample basis.

\subsection{Conditional Pointer Decoder}
To incorporate user instructions into the selection policy, text tokens are first projected and combined with the visual grid via a cross-modal input encoder consisting of bidirectional self-attention blocks. The encoding process is formalized as follows:
\begin{equation}
\tilde{\mathbf{E}}^{txt} = g_{txt} \cdot \text{LN}(\mathbf{W}_{p}\mathbf{E}^{txt}),
\label{eq:text_project}
\end{equation}
\begin{equation}
\mathbf{H} = \text{Enc}([\mathbf{X}^v; \tilde{\mathbf{E}}^{txt}])_{[1;N]} \in \mathbb{R}^{N \times D},
\label{eq:joint_encode}
\end{equation}
where $\mathbf{W}_{p} \in \mathbb{R}^{D \times D_{txt}}$ is a learned projection layer and $g_{txt} \in [0,1]$ is a scheduled gate annealed to stabilize early training. After joint encoding, only the updated tissue tokens $\mathbf{H}$ are retained as decoder memory. To support data-dependent termination, a learned stop token vector $\mathbf{s}_{stop} \in \mathbb{R}^{D}$ is appended to form the complete memory matrix:
\begin{equation}
\mathbf{M} = [\mathbf{H}; \mathbf{s}_{stop}] \in \mathbb{R}^{(N+1) \times D}.
\label{eq:memory_matrix}
\end{equation}

The autoregressive loop initializes with a learned start token $\mathbf{q}_{1} = \mathbf{s}_{start}$ and an empty log-space mask $\mathbf{Mask}_{1} \equiv \mathbf{0} \in \mathbb{R}^{N+1}$. At decoding step $t$, the decoder query layer executes causal self-attention over prior queries to generate $\mathbf{q}_t$. As shown in the step-by-step selection iterations in Fig.~\ref{fig:framework}, the pointer mechanism matches the current query $\mathbf{q}_t \in \mathbb{R}^{1 \times D}$ against the memory $\mathbf{M}$ through cross-attention logits:
\begin{equation}
\mathbf{l}_{t} = \frac{(\mathbf{q}_t \mathbf{W}_q)(\mathbf{M}\mathbf{W}_k)^\top}{\sqrt{D}} + \mathbf{Mask}_{t},
\label{eq:pointer_logits}
\end{equation}
\begin{equation}
\mathbf{p}_{t} = \text{softmax}(\mathbf{l}_{t} / \tau),
\label{eq:pointer_distribution}
\end{equation}
where $\mathbf{W}_q, \mathbf{W}_k \in \mathbb{R}^{D \times D}$ are learnable projection matrices and $\tau$ is the softmax temperature. During inference, the deterministic selection follows a hard argmax operator:
\begin{equation}
a_{t} = \arg\max_{j \in \{1, \dots, N+1\}} \mathbf{l}_{t}(j).
\label{eq:argmax_selection}
\end{equation}
To eliminate spatial redundancy and prevent duplicate choices, the token mask updates dynamically at each step:
\begin{equation}
\mathbf{Mask}_{t+1}(i) = \mathbf{Mask}_{t}(i) + (-\infty) \cdot \mathbb{I}[a_{t} = i].
\label{eq:mask_update}
\end{equation}
The selected memory slot is fed back as a column-vector query for the next token selection via a straight-through estimator (STE) to maintain differentiability:
\begin{equation}
\mathbf{q}_{t+1}^\top = \mathbf{M}^{\top}(\text{onehot}(a_{t}) - \text{sg}(\mathbf{p}_{t}^\top) + \mathbf{p}_{t}^\top),
\label{eq:ste_query}
\end{equation}
where $\text{sg}(\cdot)$ is the stop-gradient operator, ensuring that the forward pass utilizes the discrete selection $a_t$ while gradients still propagate through the soft distribution $\mathbf{p}_t$.

\subsection{Differentiable Selection via Noise-Gated Tokens}
To bypass the non-differentiable argmax during joint training, the discrete selection is rewritten as a continuous perturbation via aggregated soft scores. Let $A_{t} = \prod_{u=1}^{t-1} (1 - p_{u}(\emptyset))$ with $A_{1} = 1$ denote the alive probability that the episode has not terminated before step $t$. For each individual visual token $i$, we accumulate the alive-weighted pointer probabilities across the trajectory with a geometric depth decay base $\beta \in (0,1)$:
\begin{equation}
s_{i} = \sum_{t=1}^{T_{max}} A_{t} \cdot p_{t}(i) \cdot \beta^{t}.
\label{eq:soft_score}
\end{equation}
These raw soft scores are remapped onto a bounded range to approximate a hard top-$K$ indicator via a differentiable polarization operator:
\begin{equation}
\alpha_{i} = \text{SoftTopK}_{k,\tau}(\mathbf{s})_{i} \in [0, 1],
\label{eq:polarization}
\end{equation}
where $k = \mathbb{E}_b[K_b]$ tracks the running batch-average of the hard selection count, and $\tau$ is the shared temperature parameter controlling optimization sharpness.

Continuous variance-preserving noise gating is then enforced by mixing the original input features with isotropic Gaussian noise modulated by the polarized selection scores (refer to the Noise Gate block in Fig.~\ref{fig:framework}):
\begin{equation}
\tilde{\mathbf{x}}_{i}^v = \sqrt{\alpha_{i}}\mathbf{x}_{i}^v + \sqrt{1 - \alpha_{i}}\boldsymbol{\epsilon}_{i}, \quad \boldsymbol{\epsilon}_{i} \sim \mathcal{N}(\mathbf{0}, \mathbf{I}_{D}).
\label{eq:vp_noise}
\end{equation}
To suppress representation drift while \textbf{preventing semantic leakage}, where unselected background tokens potentially establish an information shortcut to circumvent the noise gate, a local denoiser strictly restricts cross-token feature aggregation via a diagonal attention mask:
\begin{equation}
\hat{\mathbf{X}}^v = \text{Denoise}(\tilde{\mathbf{X}}^v; \text{mask} = \text{diag}).
\label{eq:denoiser}
\end{equation}

\subsection{Joint Optimization with Length Penalty}
The refined visual block $\hat{\mathbf{X}}^v \in \mathbb{R}^{N \times D}$ undergoes a linear mapping into the linguistic domain via a multimodal bridge matrix $\mathbf{P}$. To formalize the instruction-following predictive optimization, the projected visual sequence is token-wise prepended to the text prompt tokens $\mathbf{E}^{txt}$, functioning as a multi-modal unified context layer. Let $Y = \{y_1, y_2, \dots, y_U\}$ be the target sequence of response text tokens. The language-modeling objective $\mathcal{L}_{LM}$ evaluates the conditional next-token distribution under an autoregressive maximum-likelihood formulation:
\begin{equation}
\mathcal{L}{LM}(\hat{\mathbf{X}}^v) = -\sum{u=1}^{U} \log P\left(y_u \mid [\,\mathbf{P}\hat{\mathbf{X}}^v \,;\, \mathbf{E}^{txt}\,], y_{<u}\right).
\label{eq:lm_loss}
\end{equation}
To direct the policy toward extreme visual sparsification without enforcing an inflexible predefined instance budget, we extend the optimization criterion by introducing a length penalty regularizer proportional to the compressed sample length:
\begin{equation}
\mathcal{L} = \mathcal{L}_{LM}(\hat{\mathbf{X}}^v) + \lambda \cdot \frac{K}{N},
\label{eq:total_loss}
\end{equation}
where $\lambda$ governs the trade-off between semantic density and diagnostic thoroughness. Since $K$ is determined dynamically by the initial activation of the stop action, the task error signal seamlessly backpropagates through the soft alive trajectory probability gating channels into the $\mathbf{s}_{stop}$ embedding and the pointer weights, calibrating the sequential routing parameters end-to-end.

\section{Experiments}
\label{sec:experiments}

\subsection{Experimental Setup and Implementation Details}

\textbf{Datasets and Evaluation Protocol.} 
Our model is trained and evaluated on the SlideBench (TCGA) benchmark \cite{chen2025slidechat}, reporting both overall accuracy and performance stratified across two critical categories: \textbf{Microscopy} assesses the description of low-level morphological and staining features; \textbf{Diagnosis} evaluates histology-based reasoning and clinical subtyping; \textbf{Clinical} tests the ability to retrieve and apply clinically relevant background knowledge about diseases. To assess out-of-distribution robustness, we further conduct zero-shot generalization tests on SlideBench (BCNB) and the external WSI-VQA* dataset \cite{chen2024wsi}.

\textbf{Baselines.} 
To rigorously validate our framework across diverse diagnostic dimensions, we construct a comprehensive baseline matrix. This matrix includes theoretical upper-bound uncompressed models like SlideChat \cite{chen2025slidechat}; compute-matched spatial sampling and general-purpose VLMs, including LLaVA-Med \cite{li2023llava}, Quilt-LLaVA \cite{seyfioglu2024quilt}, MedDr \cite{he2024meddr}, and the proprietary GPT-4o \cite{hurst2024gpt}.

\textbf{Training and Optimization.} 
The proposed architecture is optimized utilizing a single NVIDIA A6000 GPU via a lightweight, single-stage training protocol. Specifically, our PathSelect sequential selection module is directly integrated as an active plugin between the frozen visual encoder and the LLM of the fully pre-trained SlideChat~\cite{chen2025slidechat} base framework. To preserve the robust cross-modal alignment and generative language-following priors of this established baseline model, both the upstream slide encoder and the downstream LLM backbones are kept strictly frozen throughout the entire training duration. Parameter updates are restricted exclusively to the active networks of the newly inserted PathSelect module and its corresponding linear multimodal projection bridge $\mathbf{P}$. The policy network is optimized end-to-end using direct task error feedback backpropagated from the language modeling objective, requiring no auxiliary loss or parameter-efficient fine-tuning on the language backbone.

\begin{table}[htbp]
\centering
\caption{Main results on SlideBench (TCGA) and zero-shot generalization on SlideBench (BCNB) and WSI-VQA*.}
\label{tab:combined_results_patho}
\renewcommand{\arraystretch}{1.2} 
\setlength{\tabcolsep}{4pt} 

\begin{adjustbox}{width=\linewidth}
\begin{tabular}{lccccccc}
\toprule
\multirow{2}{*}{\textbf{Methods}} & \multirow{2}{*}{\textbf{FLOPs}} & \multicolumn{4}{c}{\textbf{SlideBench (TCGA)}} & \textbf{SlideBench} & \textbf{WSI} \\
\cmidrule(lr){3-6} 
 & & \textbf{Micros.} & \textbf{Diagnos.} & \textbf{Clinical} & \textbf{Overall} & \textbf{(BCNB)} & \textbf{(VQA*)} \\
\midrule
\rowcolor{mygray} \multicolumn{8}{c}{\textit{Upper Bound}} \\
\textcolor{gray}{SlideChat} & \textcolor{gray}{133.3T} & \textcolor{gray}{82.20} & \textcolor{gray}{72.31} & \textcolor{gray}{71.43} & \textcolor{gray}{74.81} & \textcolor{gray}{54.14} & \textcolor{gray}{60.18} \\
\midrule
Random Baseline & -- & 24.44 & 24.91 & 26.44 & 25.02 & 24.40 & 24.14 \\
GPT-4o          & -- & 62.89 & 46.69 & 66.77 & 57.91 & 41.43 & 30.41 \\
LLaVA-Med       & 1.70T & 47.34 & 32.78 & 47.96 & 42.00 & 30.10 & 26.31 \\
Quilt-LLaVA     & 1.70T & 57.76 & 35.96 & 53.07 & 48.07 & 32.19 & 44.43 \\
MedDr           & 1.70T & 73.30 & 57.78 & 74.25 & 67.70 & 33.67 & 54.36 \\
\midrule
\rowcolor{myblue} \multicolumn{8}{c}{(Pruning Rate=97.27\%, \text{LP}=2)} \\
\textbf{Ours} & 1.76T & \textbf{81.94} & \textbf{70.90} & \textbf{74.49} & \textbf{74.00} & \textbf{56.39} & \textbf{60.76} \\
\bottomrule
\end{tabular}
\end{adjustbox}
\end{table}

\subsection{Main Results}

\textbf{SlideBench Benchmark and Zero-Shot Evaluation.}
Table~\ref{tab:combined_results_patho} provides a comprehensive evaluation against both leading general-purpose models and specialized biomedical baselines. 
Under stringent capacity limits, our framework establishes a new state-of-the-art overall accuracy of $74.00\%$, substantially outperforming specialized compute-matched models such as MedDr ($+6.30\%$) and Quilt-LLaVA ($+25.93\%$), while exceeding commercial general LLM GPT-4o by $16.09\%$.

A fine-grained, subset-level analysis relative to the uncompressed SlideChat upper bound further highlights the efficacy of the proposed token selection mechanism:
\begin{itemize}
    \item \textbf{Clinical Reasoning:} Remarkably, our approach achieves $74.49\%$ accuracy, surpassing the uncompressed upper bound ($71.43\%$) by an absolute margin of $+3.06\%$, demonstrating that proper token selection can effectively mitigate redundant visual noise to enhance complex clinical reasoning.
    \item \textbf{Microscopic and Diagnostic Analysis:} On the Micros. sub-task, our model yields $81.94\%$, recovering $99.68\%$ of SlideChat's capacity (with a negligible gap of $0.26\%$). Concurrently, on the dense Diagnosis subset, it achieves $70.90\%$, outperforming the strongest baseline MedDr by $13.12\%$ and retaining $98.05\%$ of the uncompressed model's performance ($72.31\%$).
\end{itemize}
These fine-grained comparisons systematically validate that our framework accomplishes near-lossless, and in certain dimensions context-enhancing, token compression for gigapixel whole-slide image understanding.

\begin{figure}[t]
\centering
\includegraphics[width=0.98\linewidth]{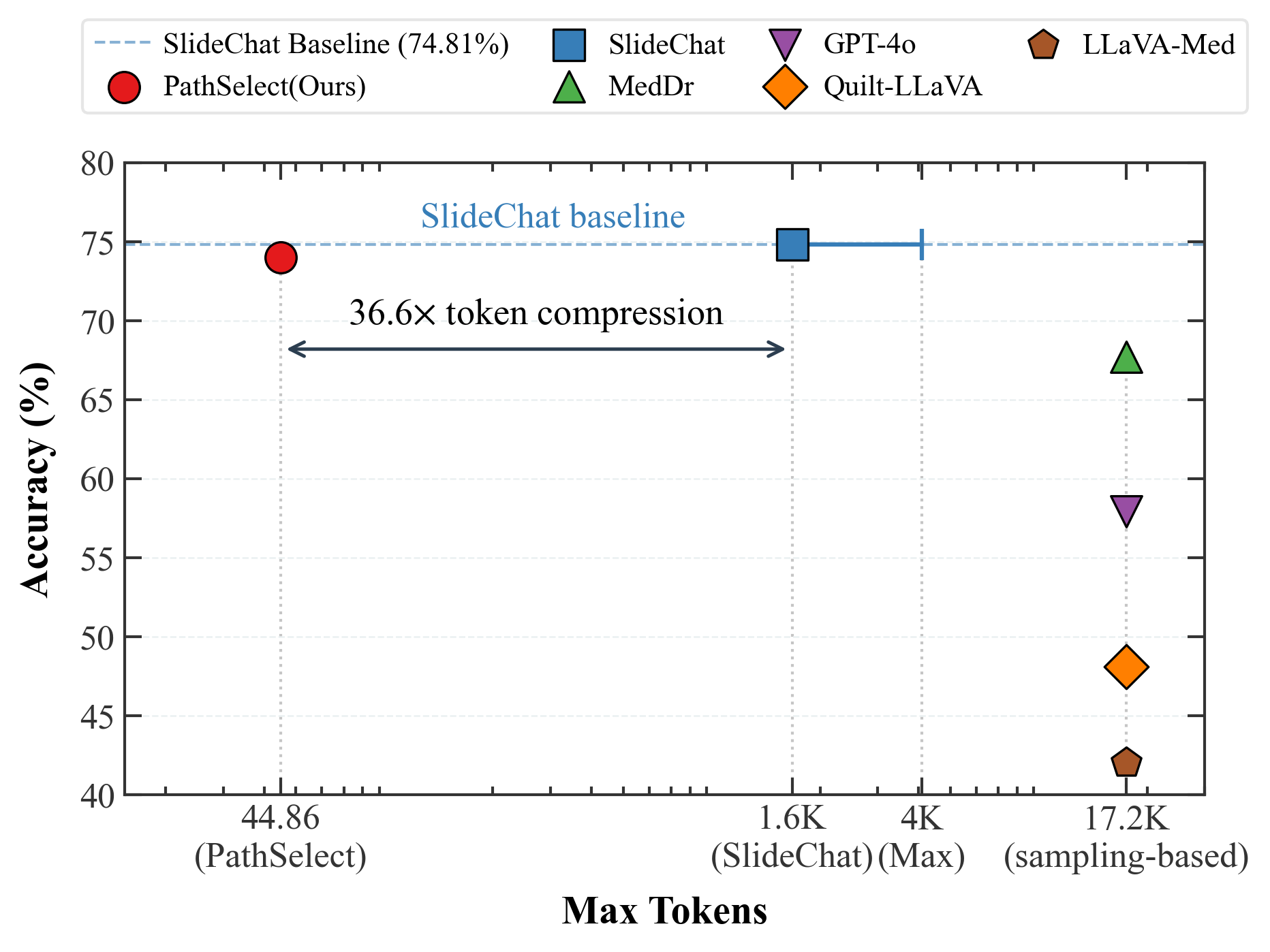}
\caption{\textbf{Performance and token efficiency comparison.} Our framework achieves competitive diagnostic accuracy using an empirical average of only $44.86$ tokens under the $LP=2.0$ configuration. This represents an approximate $36.6\times$ spatial token reduction relative to the SlideChat~\cite{chen2025slidechat} baseline average of $1,641.87$ tokens, while operating far below the maximum context ceiling of $4,096$ tokens.}
\label{fig:token_efficiency}
\end{figure}

\begin{figure}[t]
\centering
\includegraphics[width=0.98\linewidth]{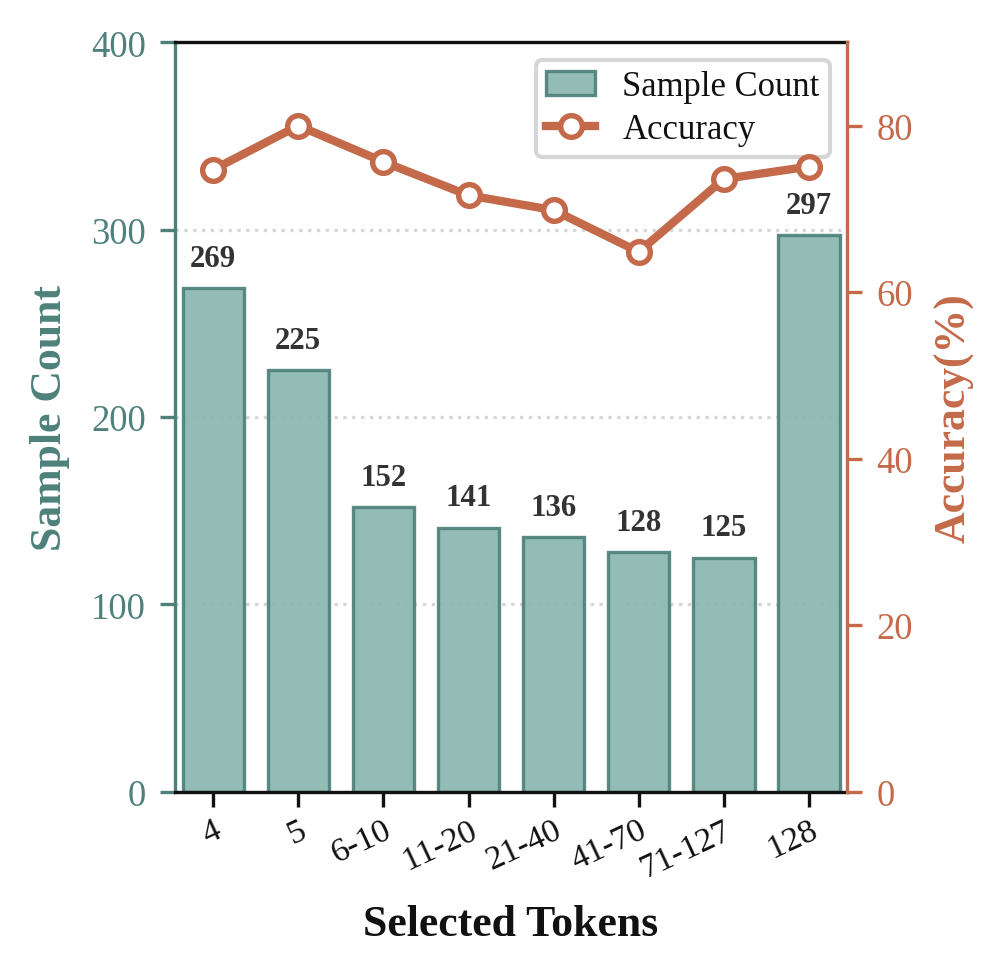}
\caption{\textbf{Accuracy stability under $LP=2.0$ training configuration.} The downstream classification accuracy (orange line) exhibits remarkable stability across all intervals of dynamically selected tokens, demonstrating consistent diagnostic robustness independent of the adaptive sample-wise compression rates.}
\label{fig:token_bins_accuracy}
\end{figure}

\subsection{Efficiency Analysis}
\label{sec:efficiency_patho}

Figure~\ref{fig:token_efficiency} quantifies the computational and token efficiency of the proposed paradigm. While the baseline SlideChat~\cite{chen2025slidechat} processes whole slide images with an empirical average sequence length of $1,641.87$ tokens (with a maximum capacity of up to $4,096$ tokens), our framework delivers near-lossless diagnostic performance with a compressed average of only $44.86$ tokens. 
By contracting the sequence length down to this highly compact scale through our text-conditioned sequential decision process, we achieve a $36.6\times$ spatial token reduction. Consequently, the quadratic computational complexity ($\mathcal{O}(L^2)$) inherent to the downstream LLM attention layers is substantially mitigated. Crucially, as formalized in Sec.~\ref{sec:methodology}, our paradigm casts token compression as an autoregressive sequential formulation rather than a static ranking framework. This data-dependent termination breaks the direct dependency of autoregressive language decoding overhead on the raw tissue patch count $N$, allowing the downstream inference costs to scale only with the intrinsic complexity of the specimen.

Figure~\ref{fig:token_bins_accuracy} further highlights the diagnostic resilience of our adaptive selection mechanism across varying token granularities. Concurrently executing under the $LP=2.0$ regularizer, our autoregressive policy network dynamically tailors the optimal token allocation $K = T^* - 1$ based on the intrinsic diagnostic difficulty of each distinct sample. 
Crucially, the classification accuracy exhibits minimal variance and remains tightly bounded around the global average across all dynamic retention intervals—even in extreme high-sparsity scenarios where only $4$ to $5$ tokens are preserved. This invariant trend empirically confirms that our pointer policy reliably captures essential diagnostic landmarks and satisfies the dynamic marginal utility without discarding critical pathological semantics.

\begin{figure}[t]
\centering
\includegraphics[width=1\linewidth]{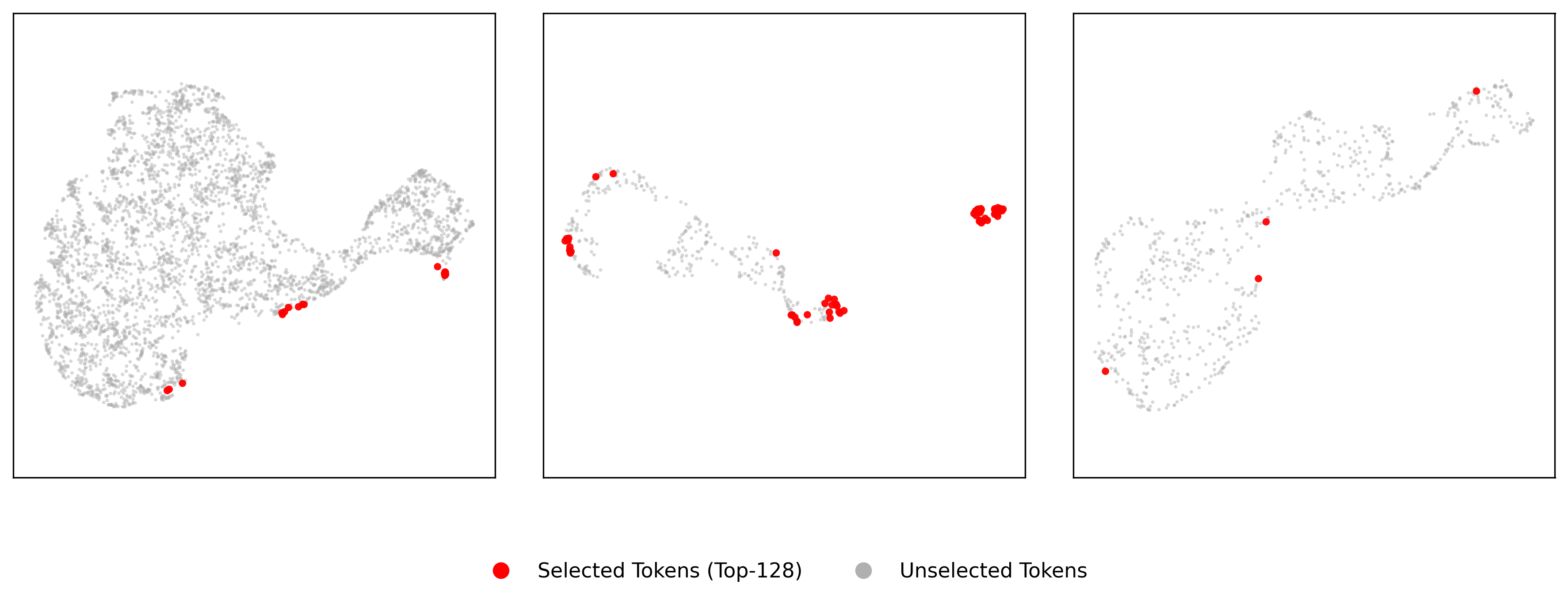}
\caption{\textbf{UMAP visualizations of the latent feature space across three diverse WSIs.} Unselected tokens cluster into dense centralized manifolds. In contrast, our autoregressive pointer uses the history of prior selections to dynamically update candidate scores, isolating complementary pathological feature outliers at the extreme topological margins.}
\label{fig:UMAP}
\end{figure}

To qualitatively evaluate the selection trajectory and verify the effects of our autoregressive pointer decoder, we project the high-dimensional patch-level visual features into a two-dimensional embedding space via UMAP, as illustrated in Figure~\ref{fig:UMAP}. The vast majority of unselected tokens cluster into dense, centralized manifolds, which correspond to redundant stromal regions or healthy background tissues. 
Conversely, the sparse tokens selected by our model consistently occupy the extreme topological margins of the feature distribution. This distinctive distribution empirically cross-references the sequential routing capability defined in Eq.~\eqref{eq:autoregressive_policy}. 
Unlike static ranking frameworks that evaluate each candidate token in isolation, our autoregressive pointer updates candidate selection scores conditioned on the historical trajectory of prior choices. 
By incorporating this dynamic sequential dependency, the selection policy naturally suppresses immediate spatial and semantic redundancy, sequentially routing complementary pathological outliers that dictate slide-level clinical reasoning, as mathematically formulated in Eq.~\eqref{eq:soft_score}..

\subsection{Ablation Studies}
\label{sec:ablation_patho}

We conduct rigorous ablation experiments to validate the architectural necessity of the core components within our learning framework. All variants are evaluated under identical training configurations, and the quantitative comparisons are summarized in Table~\ref{tab:ablation_fixed_width} and Table~\ref{tab:length_penalty_ablation}.

\begin{table}[ht]
\centering
\caption{Ablation study on model architectural variants. Results are reported in terms of average accuracy (\%).}
\label{tab:ablation_fixed_width}
\small
\renewcommand{\arraystretch}{1.15} 
\begin{tabular}{p{5.1cm} | c}
\toprule
\textbf{Variant Configuration} & \textbf{Avg. Acc. (\%)} \\
\midrule
\multicolumn{2}{l}{\textbf{(a) Attention in Denoiser}} \\
\quad Global attention & 73.25 \\
\quad Diagonal attention$^\dagger$ & 74.00 \\
\midrule
\multicolumn{2}{l}{\textbf{(b) Scorer Aggregation}} \\
\quad Last-step pointer only & 73.31\\
\quad Uniform sum ($A_t \equiv 1, \beta = 1$) & 72.91 \\
\quad Alive-weighted only ($\beta = 1$) & 73.16 \\
\quad $A_t \cdot p_t(i) \cdot \beta^{t\,\dagger}$ & 74.00 \\
\bottomrule
\end{tabular}
\end{table}

\textbf{Attention Mechanisms in Denoiser.} 
As reported in Table~\ref{tab:ablation_fixed_width}(a), replacing the proposed diagonal attention mask with a full global self-attention mechanism yields a performance degradation from 74.00\% to 73.25\%. This performance drop validates our hypothesis regarding semantic leakage formulated in Sec.~\ref{sec:methodology}. When global attention is permitted within the denoiser, unselected or low-scoring background tokens can aggregate crucial diagnostic contexts from high-score informative tokens. Consequently, the downstream LLM circumvents the variance-preserving noise bottleneck defined in Eq.~\eqref{eq:vp_noise} by retrieving masked semantics from the perturbed channels, which disrupts the optimization consistency between the differentiable training path and the deterministic inference path.

\textbf{Scorer Aggregation Strategies.} 
Table~\ref{tab:ablation_fixed_width}(b) evaluates four trajectory aggregation formulations for constructing the soft score vector $\mathbf{s}$ in Eq.~\eqref{eq:soft_score}. 
(i) Utilizing the pointer distribution from the final step alone (\textit{Last-step pointer only}) degrades the average accuracy to 73.31\%, demonstrating that a single-step decision fails to encapsulate the historical cumulative marginal utility of the sequence. 
(ii) A naive uniform summation (\textit{Uniform sum}, where $A_t \equiv 1, \beta = 1$) reduces accuracy to 72.91\%, as it fails to penalize tracking errors across extended trajectories and masks the temporal dynamics of the pointer. 
(iii) Removing the geometric depth decay while retaining the termination probability (\textit{Alive-weighted only}, $\beta = 1$) yields 73.16\%, which verifies the necessity of temporal discounting in long-context decision-making. 
(iv) The complete formulation incorporating both the alive probability $A_t$ and the geometric depth decay $\beta^t$ as specified in Eq.~\eqref{eq:soft_score} achieves the optimal accuracy of 74.00\%, confirming that joint modeling of execution state and sequential urgency provides the most effective gradient path for the policy network.

\begin{table*}[t]
\centering
\caption{Ablation study on the length penalty functions and compression metrics on SlideBench-TCGA dataset. Each classification metric reports the mean and standard deviation over 2 or 3 independent runs.}
\label{tab:length_penalty_ablation}
\small
\resizebox{\linewidth}{!}{%
\setlength{\tabcolsep}{8pt} 
\begin{tabular}{l c c c cccc}
\toprule
\multicolumn{2}{c}{\textbf{Hyperparameters}} & \multicolumn{2}{c}{\textbf{Token Metrics}} & \multicolumn{4}{c}{\textbf{SlideBench-TCGA Performance}} \\
\cmidrule(r){1-2} \cmidrule(lr){3-4} \cmidrule(l){5-8} 
Method / Function & $LP$ & Avg. Tokens & Max. Tokens & Microscopy (\%) & Diagnosis (\%) & Clinical (\%) & Average (\%) \\
\midrule
Length Penalty Sweep & 0.0 & 66.07 & 128 & 81.55 $\pm$ 0.56 & 70.59 $\pm$ 0.57 & 72.96 $\pm$ 0.72 & 73.59 $\pm$ 0.48 \\ \addlinespace[0.3em]
($K=128$)            & 0.5 & 51.80 & 128 & 81.42 $\pm$ 0.37 & 70.69 $\pm$ 0.57 & 72.96 $\pm$ 0.72 & 73.62 $\pm$ 0.43 \\ \addlinespace[0.3em]
                     & 1.0 & 48.09 & 128 & 81.42 $\pm$ 0.37 & 70.80 $\pm$ 0.00 & 73.47 $\pm$ 1.44 & 73.72 $\pm$ 0.19 \\ \addlinespace[0.3em]
                     & 2.0 & 45.84 & 128 & 81.94 $\pm$ 0.00 & 70.69 $\pm$ 0.29 & 73.47 $\pm$ 1.44 & 73.80 $\pm$ 0.29 \\ \addlinespace[0.3em]
                     & 4.0 & 43.25 & 128 & 81.28 $\pm$ 0.56 & 71.05 $\pm$ 0.35 & 72.45 $\pm$ 0.00 & 73.79 $\pm$ 0.39 \\ \addlinespace[0.3em]
                     & 6.0 & 39.54 & 128 & 81.56 $\pm$ 0.18 & 70.59 $\pm$ 0.15 & 73.47 $\pm$ 0.00 & 73.62 $\pm$ 0.15 \\
\midrule 
Target Budget Sweep  & 2.0 & 42.91 & 64  & 80.89            & 71.30            & 73.47            & 73.93            \\ \addlinespace[0.3em]
($LP=2.0$)           & 2.0 & 45.84 & 128 & 81.94            & 70.69            & 73.47            & 73.80            \\ \addlinespace[0.3em]
                     & 2.0 & 61.43 & 256 & 82.20            & 70.09            & 71.43            & 73.32            \\
\bottomrule
\end{tabular}%
}
\end{table*}

\textbf{Sensitivity Analysis of Length Penalty and Target Budgets.}
To explore the trade-off governed by the length penalty coefficient $\lambda$ defined in Eq.~\eqref{eq:total_loss}, we execute a systematic hyperparameter sweep as summarized in Table~\ref{tab:length_penalty_ablation}. 
Under a maximum constraint of $K=128$, escalating $LP$ ($\lambda$) from $0.0$ to $2.0$ induces a monotonic contraction in the average retained token count from $66.07$ to $45.84$, while simultaneously facilitating a progressive increase in downstream average accuracy from $73.59\%$ to $73.80\%$. 
This positive correlation indicates that optimizing the length regularizer actively purges non-informative visual regions, mitigating token redundancy before linguistic processing. 
However, when $\lambda$ is aggressively scaled beyond $4.0$, the average token retention falls below $40$, leading to a performance plateau or slight decay due to the inadvertent filtering of critical visual diagnostics.

Furthermore, we evaluate variations in the target budget capacity. When fixing $LP=2.0$, compressing the sequence down to a maximum of 64 tokens yields an optimal accuracy of 73.93\% with merely 42.91 average tokens. Conversely, inflating the maximum allocation ceiling to 256 tokens causes a noticeable performance drop to 73.32\%, despite expanding the sequence length. This trend empirically corroborates that unconstrained token capacities introduce significant representation noise to the frozen LLM, highlighting the dual advantages of our dynamic sequential selection paradigm in optimizing both computational efficiency and diagnostic precision.

\section{Conclusion}
\label{sec:conclusion}

In this work, we reformulate WSI token pruning as a \textbf{sequential selection process} by embedding our PathSelect plugin into the pre-trained SlideChat architecture. By decoupling VP noise gating during training from deterministic Hard Top-$K$ token pruning at inference, the PathSelect training module elegantly resolves the non-differentiability of discrete sequential decisions via a lightweight, single-stage training protocol on a single GPU. Driven by a highly efficient empirical average of $44.86$ tokens, our approach yields $74.00\%$ overall accuracy on SlideBench (TCGA) and distinctly improves cross-modal diagnostic alignment without discarding sparse pathological evidence.

\bibliographystyle{unsrt}
\bibliography{main}

\end{document}